\documentclass[lettersize,journal]{IEEEtran}
\usepackage{mathtools, mathrsfs, amsthm, bm}
\usepackage{newtxtext, newtxmath}

\usepackage{algorithm}
\usepackage{algpseudocode}
\usepackage{booktabs}
\usepackage{multirow}
\usepackage{tabularx}
\usepackage{array}
\usepackage{colortbl}

\usepackage[table]{xcolor}

\usepackage{graphicx}
\usepackage{float}
\usepackage[caption=false,font=normalsize,labelfont=sf,textfont=sf]{subfig}

\usepackage{hyperref}
\usepackage{orcidlink}
\usepackage{textcomp}
\usepackage{url}
\usepackage{verbatim}
\usepackage{stfloats}
\usepackage{balance}
\def\BibTeX{{\rm B\kern-.05em{\sc i\kern-.025em b}\kern-.08em
		T\kern-.1667em\lower.7ex\hbox{E}\kern-.125emX}}
\begin{document}

\title{MoRPI-PINN: A Physics-Informed Framework for Mobile Robot Pure Inertial Navigation}

\author{
	Arup Kumar Sahoo %\orcidlink{0000-0003-4515-7434} 
	\thanks{ A. K. S.  is supported in part by the Maurice Hatter Foundation. (Corresponding author: Arup Kumar Sahoo)}
	and Itzik Klein %\orcidlink{0000-0001-7846-0654 }
	\thanks{ The authors are with the Autonomous Navigation and Sensor Fusion Lab (ANSFL), the Hatter Department of Marine Technologies,
		Charney School of Marine Sciences, University of Haifa, Haifa 3498838,
		Israel. (e-mail: asahoo@campus.haifa.ac.il, kitzik@univ.haifa.ac.il } }

%\markboth{Journal of \LaTeX\ Class Files,~Vol.~x, No.~x, June~2025}%
%{How to Use the IEEEtran \LaTeX \ Templates}

\maketitle

\begin{abstract}
A fundamental requirement for full autonomy in mobile robots is accurate navigation even in situations where satellite navigation or cameras are unavailable. In such practical situations, relying only on inertial sensors will result in navigation solution drift due to the sensors' inherent noise and error terms. One of the emerging solutions to mitigate drift is to maneuver the robot in a snake-like slithering motion to increase the inertial signal to noise ratio allowing the regression of the mobile robot position. In this work, we propose MoRPI-PINN as a physics-informed neural network framework for accurate inertial-based mobile robot navigation. By embedding physical laws and constraints into the training process, MoRPI-PINN is capable of providing an accurate and robust navigation solution. Using real-world experiments, we show accuracy improvements of over 85\% compared to other approaches. MoRPI-PINN is a lightweight approach that can be implemented even on edge devices and used in any typical mobile robot application.
\end{abstract}


\begin{table}[htbp]
	\centering
	{\renewcommand{\thetable}{} % Remove number
		\caption{\textbf{List of Abbreviations}} % No number shown
		%\label{tab:abbreviations}
	}
\addtocounter{table}{-1}
	\renewcommand{\arraystretch}{1.2}
	\begin{tabular}{@{}ll@{}} % remove left/right padding
		\toprule
		\textbf{Abbreviation} & \textbf{Definition} \\
		\midrule
		\multicolumn{2}{@{}l}{\textbf{Navigation}} \\
		INS    & Inertial Navigation System \\
		IMU    & Inertial Measurement Unit \\
		GNSS   & Global Navigation Satellite System \\
		RTK    & Real-Time Kinematic \\
		NED    & North-East-Down (Coordinate Frame) \\
		ECEF   & Earth-Centered Earth-Fixed \\
		\addlinespace
		\multicolumn{2}{@{}l}{\textbf{Neural Networks}} \\
		PINN   & Physics-informed Neural Network \\
		AD     & Automatic Differentiation \\
		GT     & Ground Truth \\
		\addlinespace
		\multicolumn{2}{@{}l}{\textbf{Error Metrics}} \\
		ATE    & Absolute Trajectory Error \\
		MATE   & Mean Absolute Trajectory Error \\
		MSE    & Mean Squared Error \\
		RMSE   & Root Mean Squared Error \\
		NRMSE  & Normalized Root Mean Squared Error \\
		\bottomrule
	\end{tabular}
\end{table}

\section{Introduction}
\noindent \IEEEPARstart{T}{he} study of mobile robots is crucial for modern industries and services, as they are deployed across various fields. It includes logistics, transportation, agriculture, healthcare, military operations, and data collection in hazardous environments \cite{gao2018review, shalal2013review, hurwitz2024deep}, and \cite{rao2025vibration}. In recent years, the market demand of mobile robots have been significantly increased, because of increase in application areas, advances in robot technology along with the cost reductions in electronic sensors and devices. As a result, around the globe, various companies are developing mobile robots to meet these demands and explore new markets.

\noindent Navigation plays a fundamental role in mobile robot design, enabling precise determination of position and orientation across diverse and often challenging environments. 
%It may be classified into indoor or outdoor navigation. Furthermore, the indoor navigation approaches are generally categorized into three main groups \cite{desouza2002vision} (a) map-based navigation, (b) map-building-based navigation, and (c) mapless navigation. Unlike indoor navigation, outdoor settings often lack comprehensive maps, compelling robots to adaptively perceive in real time. Outdoor navigation can be further divided into (a) structured environments, and (b) unstructured environments. However, these navigation, depends 
While operating outdoors, the navigation task depends on a variety of sensors, including cameras~\cite{desouza2002vision}, LiDAR~\cite{cheng2018mobile}, Sonar~\cite{tsalatsanis2007mobile}, global navigation satellite systems (GNSS)~\cite{suzuki2014autonomous}, inertial navigation system (INS)~\cite{maaref2002sensor}, and odometers~\cite{nemec2020estimation}. 
In indoors or tunnels the GNSS signals are unavailable while vision methods suffer from poor lighting conditions or featureless environments. In such real-world scenarios, the navigation solution depends only on the inertial sensor readings in a process known as pure inertial navigation~\cite{titterton2004strapdown, farrell2007gnss}.
%But sometimes it struggle in indoor environments, tunnels, under water and low-light scenarios where vision-based methods or GNSS signals are unreliable. Nonetheless, in practical situations only the inertial readings are available.

%\noindent Inertial measurement units (IMUs), composed of tri-axial accelerometers and gyroscopes~\cite{klein2025pedestrian}, are widely used owing to their small size, low cost, high sampling rates, and ease of integration. However, in the real world scenario, mobile robots mounted with IMUs often generate noisy readings when moving at a nearly constant velocities. Despite their advantages, inertial navigation is prone to drift over time, due to the accumulation of noise and integration errors. It results in degrading the  positional accuracy. It is also less effective under adverse environmental conditions. 

\noindent There, the navigation solution drifts due to errors and noise in the inertial measurement. To cope with the inertial drift Shurin and Klein~\cite{shurin2020qdr} proposed the quadrotor dead-reckoning approach. Inspired by the natural movement of pedestrians~\cite{klein2025pedestrian, liu2025novel}, they enforced periodic motion on quadrotors and designed a model-based approach for positioning. Later, similar model-based approaches were applied on mobile robots~\cite{etzion2023morpi}. With the emergence of deep-learning approaches in the navigation field \cite{cohen2024inertial, chen2024deep}, neural network solutions were designed for quadrotors \cite{shurinquadnet} and mobile robots \cite{etzion2025snake}.

%\noindent Recently, biologically inspired approaches have influenced navigation strategies. A novel technique by Shurin et al.~\cite{shurin2020qdr} mimics serpentine movement to improve maneuverability, robustness, and sensor data quality through signal-to-noise ratios for quadrotors. Etzion and Klein  proposed a novel Mobile Robot Pure Inertial Navigation (MoRPI) framework, which leverages periodic motion and employs an empirical formula to estimate step length of a wheeled robot. However, in subsequent works, they have implemented deep learning models in order to improve the results \cite{shurinquadnet, etzion2025snake}.  Additionally, in this experiment \cite{etzion2025snake}, multiple IMUs were mounted on a mobile robot exhibiting snake-like slithering motion to estimate its dead reckoning position. In a recent work, Cohen and Klein \cite{cohen2024inertial} have been employed data-driven methods in autonomous underwater vehicles (AUV) with promising outcomes. In a followup work, Yampolsky and Klein \cite{yampolsky2025dcnet} modeled another data-driven methods  for AUV related tasks. Moreover, the applications of supervised deep learning models demonstrated effective performance across various unmanned vehicles listed above such as quadrotors, wheeled robots and AUVs. Nevertheless, it depend on large labeled datasets and often yield physically inconsistent predictions. 

\noindent Recent advances in scientific machine learning, offers promising solutions to such challenges and uncovers novel scientific phenomena, particularly through physics-informed neural networks (PINNs). Unlike conventional deep neural networks (DNNs),  PINNs embed physical laws typically expressed as differential equations (DEs) directly into the objetive function of neural network as a residual loss \cite{raissi2019physics}. This integration ensures that model outputs adhere to governing physical principles, enhancing accuracy and interpretability while reducing reliance on extensive labeled data. PINNs leverage parallel computing and autometic differentiation, in order to address the computational inefficiencies of repeatedly solving DEs. In past few years, PINNs have demonstrated remarkable success across scientific and engineering domains, including anomaly detection \cite{dwivedi2023dynamopmu, lim2025physics},
fluid mechanics \cite{kumar2023physics}, solid mechanics \cite{sahoo2024unsupervised}, remote sensing \cite{guo2024analysis}, robot manipulation \cite{gu2024physics, park2024state}, and beyond.  These advances highlight PINNs as a powerful tool for modeling complex dynamical systems, particularly where traditional methods struggle due to noisy data. 
% In a pioneer work, PINNs have been integrated with model predictive control to improve trajectory tracking for robots and autonomous vehicles, achieving higher accuracy and faster response compared to traditional methods \cite{liu2024research}. Sanyal and Roy \cite{sanyal2023ramp} used same approach to enable quadrotor trajectory tracking under uncertain dynamic disturbances. Nicodemus et al. \cite{nicodemus2022physics} applied PINNs to the trajectory tracking problem in multi-link manipulators. By introducing an over-sampling strategy and a zero-holding assumption, they effectively reduced the dimensionality of the input space, enabling the PINN-based controller to closely approximate the system dynamics.
   In the domain of navigation and control, Xu et al. \cite{xu2022physics} developed a PINN-based model for the prediction in sway and surge velocities, as well as rotational speed in the unmanned surface vehicles. In a recent work, Park and Lee \cite{park2024state} employed PINN algorithm with error tracking control to estimate the speed of the leader robot.

\noindent Building on the PINN advancements, this paper introduces MoRPI-PINN, a mobile robot pure inertial framework for accurate navigation. It is hypothesized that PINNs can effectively guide the model using partial physical knowledge in the form of ordinary or partial differential equations and limited data. PINNs offer a unified framework for such scenarios by embedding known physical laws (e.g., 2D-INS equations here) directly into neural network training. This is achieved through automatic differentiation enforcing the governing differential equations as soft constraints in the loss function. In inertial navigation, MoRPI-PINN model learn the system’s trajectory by balancing physical consistency and data fidelity. This makes them particularly suitable for navigation tasks with limited and/or noisy sensor data. To further increase the influence of our MoRPI-PINN framework, we constrain the robot to move in a snake-like slithering motion, as this motion has already proven to yield an increased inertial signal to noise ratio allowing regression of the mobile robot's position, even in rough terrain.

\noindent The contributions of this paper are:
\begin{itemize}
	\item	Development of the MoRPI-PINN framework to cope with real-world scenarios of pure inertial navigation for mobile robots operating in various scenarios. 
		
	\item	Integration of the governing physics of 2D-INS equations of motion with sparse sensor data, relying only on a single trajectory, during the training process of the network.
\end{itemize}
 \noindent We demonstrate that by embedding the physical laws and constraints of 2D-INS equations of motion into the training process, MoRPI-PINN provides an accurate and robust navigation solution for a mobile robot. Using real-world experiments on a mobile robot equipped with IMU and RTK-GNSS, we present an 85\% improvement over other model-based and data-driven methods.

\noindent The rest of the paper can be outlined as follows: Section \ref{sec2} presents the INS equation and various model based solutions. Section \ref{sec3} gives our proposed approach, detailing the formulation of PINNs algorithm and model architechture. Section \ref{sec4} describes our experimental setup and presents comprehensive results, with analyses of the data. Finally, the conclusions are drawn in Section \ref{sec5}.

\section{Model-Based Approaches} \label{sec2}
\noindent This section provides a brief overview of 2D-INS model-based solutions  and MoRPINet prediction, that has been later used as baselines for comparison with our proposed MoRPI-PINN approach.
\subsection{Inertial Navigation System}
\noindent An inertial navigation system provides a comprehensive navigation solution, encompassing position, velocity, and orientation. Moreover, the fundamental INS equations of motion are often formulated within a navigation frame (n-frame) using north-east-down (NED) coordinates. But for practical mobile robot navigation, a more convenient local frame (l-frame) is typically employed. This l-frame is defined at the starting location of robot, with its axes aligned to the north, east, and down directions.
The INS equations of motion are given by \cite{etzion2023morpi}:
\begin{align} 
	\dot{\mathbf{p}}^n &= \mathbf{v}^n, \label{eq:position} \\ 
	\dot{\mathbf{v}}^n &= \mathbf{C}_b^n \mathbf{f}_{ib}^b + \mathbf{g}^n, \label{eq:velocity} \\ 
	\dot{\mathbf{C}}_b^n &= \mathbf{C}_b^n \mathbf{\Omega}_{ib}^b. \label{eq_ins}
\end{align}

\noindent Where in \eqref{eq:position}--\eqref{eq_ins}, \(\mathbf{p}^n\) denotes the position vector in the l-frame, \(\mathbf{v}^n\) represents the velocity vector in the same frame, and \(\mathbf{g}^n\) is the gravity vector, which is assumed to be constant throughout the trajectory. The matrix \(\mathbf{C}_b^n\) is an orthonormal rotation matrix that transforms vectors from the body frame (b-frame) to the n-frame. The specific force \(\mathbf{f}_{ib}^b\) is measured by the accelerometer in the b-frame, while \(\mathbf{\Omega}_{ib}^b\) is a skew-symmetric matrix representing the angular rate measured by the gyroscope.

\noindent The accelerometer measures the specific force in the b-frame, represented by the vector:
\begin{equation}
	\mathbf{f}_{ib}^b =
	\begin{bmatrix}
		f_x & f_y & f_z
	\end{bmatrix}^T,
\end{equation}
where \( f_x, f_y, f_z \) are the accelerometer readings along the b-frame axes.

\noindent Similarly, the gyroscope measures the angular velocity in the b-frame as:
\begin{equation}
	\mathbf{\omega}_{ib}^b =
	\begin{bmatrix}
		\omega_x & \omega_y & \omega_z
	\end{bmatrix}^T.
\end{equation}
Here, \( \omega_x, \omega_y, \omega_z \) are the components of the angular rate vector. The subscript \( ib \) indicates measurement relative to the l-frame, while the superscript \( b \) denotes resolution in the b-frame. The skew-symmetric angular rate matrix \( \mathbf{\Omega}_{ib}^b \) is defined as:
\begin{equation}
	\mathbf{\Omega}_{ib}^b =
	\begin{bmatrix}
		0 & -\omega_z & \omega_y \\
		\omega_z & 0 & -\omega_x \\
		-\omega_y & \omega_x & 0
	\end{bmatrix},
\end{equation}
where \( \omega_x, \omega_y, \omega_z \) are the components of the angular velocity vector \( \mathbf{\omega}_{ib}^b \).

\subsection{Wheeled Robot Motion Model}
\noindent For wheeled robot motion, it is commonly assumed that the roll and pitch angles are negligible, with movement primarily occurring in the \( x \)-\( y \) plane. Under this assumption, the body-to-navigation frame transformation depends solely on the yaw angle \( \psi \), and is given by \cite{groves2013book}:
\begin{equation}
	\mathbf{C}_b^n =
	\begin{bmatrix}
		\cos \psi & -\sin \psi & 0 \\
		\sin \psi & \cos \psi & 0 \\
		0 & 0 & 1
	\end{bmatrix}.
	\label{eq_5}
\end{equation}
\noindent Substituting ~(\ref{eq_5}) into ~(\ref{eq:velocity}) shows that \( f_z \) has no influence on position or velocity in the horizontal plane. Thus, it can be excluded from inertial calculations. Additionally, since only the yaw angle is relevant, the gyroscope components \( \omega_x \) and \( \omega_y \) are neglected, and only \( \omega_z \) is retained. Furthermore, due to the use of low-cost inertial sensors and the relatively short time intervals involved, both the Earth's rotation rate and transport rate are neglected in ~(\ref{eq:velocity})--(\ref{eq_ins}).

\subsection{MoRPINet}

\noindent MoRPINet is a state-of-the-art data driven model specifically designed for pure inertial positioning approach of mobile robot. It leverages the serpentine dynamic motion of mobile robots and follows a three-phase process for dead-reckoning position updates. It can be listed as:

\begin{itemize}
\item \textbf{Distance Estimation:} A neural network architecture viz. D-Net is introduced for distance regression. It combines one-dimensional convolutional layers (1D-CNN) with a fully connected output layer to estimate distance using only inertial sensor data.

\item \textbf{Heading Estimation:} The widely used Madgwick filter is employed to determine heading based on inertial measurements. Moreover, the Madgwick filter depends on two parts: (1) orientation from angular rate, and (2) orientation from vector observations.

\item \textbf{Position Update:} Finally, the position is updated through dead reckoning, utilizing the distance and heading estimates obtained from the previous steps. The update has been performed using the following equations:
\begin{align}
	x_{k+1} &= x_k + s_{\text{DNet},k} \cos \psi_k \label{eq:x_update} \\
	y_{k+1} &= y_k + s_{\text{DNet},k} \sin \psi_k \label{eq:y_update}
\end{align}
where \( s_{\text{DNet},k} \) denotes the distance estimated by D-Net at time step \( k \), and \( \psi_k \) is the heading angle estimated by Madgwick filter. 
\end{itemize}
It is worth mentioning that the MoRPINet does not provide velocity information.

\section{Proposed MoRPI-PINN Framework} \label{sec3}
\noindent In this work, we propose MoRPI-PINN, a novel pure inertial positioning framework designed for wheeled robots moving in periodic motion. MoRPI-PINN leverages the enforced maneuverability of such robots to learn the periodic trajectory patterns. A designed error-state observer, incorporating the 2D-INS equations of motion as physical constraints, is integrated into the model to enhance prediction accuracy. The MoRPI-PINN framework fuses these known physical constrains with sensor data to accurately track the trajectory of the mobile robot.

\subsection{MoRPI-PINN Loss Function}
\noindent The construction and optimization of the PINN loss function is critical for the success of the task at hand. In the training process of our MoRPI-PINN approach, the primary objective is to optimize the network parameters (weights and biases) to satisfy both the observed inertial sensor data and the underlying physical constraints enforced by the 2D-INS dynamics.

\noindent Our proposed neural network is parameterized by $\theta$, which represents all trainable weights and biases of the network. During training, $\theta$ is optimized to minimize the total loss function, denoted as $\mathscr{L}$, integrates both a supervised loss $(\mathscr{L}_{\text{data}})$ from accelerometers and gyroscopes measurements, and an unsupervised physics loss $(\mathscr{L}_{\text{phys}})$ derived from the 2D-INS equations of motion, along with initial condition loss $(\mathscr{L}_{\text{init}})$. Each of the sub-loss terms is described in detail below.

\begin{itemize}
	\item {\bfseries Data Loss:} Penalizes the discrepancy between the predicted states (position and velocity) and their GT counterparts. The data loss is calculated as the mean squared error (MSE) between the network’s predictions and the GT values of the mobile robot 2D position and velocity vectors. 
	\begin{align}
		\mathscr{L}_{\text{data}}= \frac{1}{N_{\text{data}}} \sum_{i=1}^{N_{\text{data}}} \Bigg(  \| \mathbf{p}_{i} - \hat{\mathbf{p}}_{i} \|^2 + \| \mathbf{v}_{i} - \hat{\mathbf{v}}_{i} \|^2  \Bigg),
		\label{eq:data}
	\end{align}
	where $\mathbf{p}_i$ and $\mathbf{v}_i$ represent the GT values of 2D position and velocity vectors of the mobile robot at data point $ i $, respectively, while $\hat{\mathbf{p}}_i$ and $\hat{\mathbf{v}}_i$ denote the corresponding predicted vectors obtained from the network. The term $N_{\text{data}}$ refers to the number of supervised training samples for which GT measurements are available.

	\item \textbf{Initial Condition Loss:} In order to ensure accurate trajectory estimation at the beginning of each input window, an initial condition loss is introduced. It penalizes discrepancies between the predicted initial position and its known initial values. It is computed as follows:
	\begin{equation}
		\mathscr{L}_{\text{init}} = \frac{1}{N_{\text{ic}}} \sum_{i=1}^{N_{\text{ic}}} \Bigg(  \|\hat{\mathbf{p}}_i - \mathbf{p}_0 \|^2  \Bigg),
		\label{eq:ic}
	\end{equation}
	where $ \mathbf{p}_0 $ is the initial position vector and $N_{\text{ic}}$ is the number of samples used for enforcing the initial condition loss. Here, we set \( N_{\text{ic}} = 200 \), corresponding to 200 initial condition points used during training. These points are generated by repeating the first observed sensor readings from the training data. 

	\item \textbf{Physics Loss}:
	In addition to the supervised learning losses, a physics-informed loss is incorporated into the model to ensure that the predictions of network are consistent with the underlying 2D-INS dynamics. The physics-informed residual loss function is calculated as:
	\begin{align}
		&\mathscr{L}_{\text{phys}} = \frac{1}{N_{\text{phys}}} \sum_{i=1}^{N_{\text{phys}}}  \Bigg( \left\| \frac{d \hat{\mathbf{p}}^n(t_i; \theta)}{dt} - \hat{\mathbf{v}}^n(t_i; \theta) \right\|^2 \notag \\
		&\quad + \left\| \frac{d \hat{\mathbf{v}}^n(t_i; \theta)}{dt} - \left( \hat{\mathbf{C}}_b^n(t_i; \theta) {\mathbf{f}}_{ib}^b(t_i) + {\mathbf{g}}^n(t_i) \right) \right\|^2 \notag \\
		&\quad + \left\| \frac{d \hat{\mathbf{C}}_b^n(t_i; \theta)}{dt} - \hat{\mathbf{C}}_b^n(t_i; \theta){\mathbf{\Omega}}_{ib}^b(t_i) \right\|_F^2 \Bigg),
	\end{align}
	where $ N_{\text{phys}} $ is number of collocation points used to enforce the underlying physical laws through the physics loss. Here, we generate a set of \( N_{\text{phys}} =2000 \) collocation points, which are sampled uniformly across the time domain and randomly within the input sensor space. It ensures that the physics loss is evaluated not only at observed data points but also throughout the broader spatiotemporal domain; for generalization and consistency with the 2D-INS dynamics. Moreover, $\theta$ represents the set of trainable parameters of the neural network, and the notation $\| \cdot \|_F$ represents the Frobenius norm for matrices. This loss computes the discrepancy between the network predicted dynamics and the true time derivative of the states, based on the physical model given in (\ref{eq:position})--(\ref{eq_ins}).

		\noindent The physics loss can be expressed in explicitly as:
	\begin{align}
		&	\mathscr{L}_{\text{phys}} = \frac{1}{N_{\text{phys}}} \sum_{i=1}^{N_{\text{phys}}} \Bigg[
		\left\| \frac{d}{dt} 
		\begin{bmatrix}
			\hat{x}(t_i) \\
			\hat{y}(t_i)
		\end{bmatrix}
		- 
		\begin{bmatrix}
			\hat{v}_x(t_i) \\
			\hat{v}_y(t_i)
		\end{bmatrix}
		\right\|^2 \notag \\
		&+ \left\| \frac{d}{dt} 
		\begin{bmatrix}
			\hat{v}_x(t_i) \\
			\hat{v}_y(t_i)
		\end{bmatrix}
		-
		\left( 
		\mathbf{C}_b^n(\hat{\psi}(t_i)) 
		\begin{bmatrix}
		{f}_x(t_i) \\
		{f}_y(t_i)
		\end{bmatrix}
		+
		\begin{bmatrix}
			g_x \\
			g_y
		\end{bmatrix}
		\right)
		\right\|^2 \notag \\
		&+ \left( \frac{d\hat{\psi}(t_i)}{dt} -{\omega}_z(t_i) \right)^2
		\Bigg],
		\label{eq:phyloss}
	\end{align}
	where 	$ \hat{x}(t_i) $ and $ \hat{y}(t_i) $ are predicted position components, respectively, 
$\hat{v}_x(t_i) $ and $ \hat{v}_y(t_i) $ are predicted velocity components, respectively, and $ \hat{\psi}(t_i) $	is predicted yaw angle. The 2D body to navigation rotation matrix \(\mathbf{C}_b^n(\hat{\psi}) \) is expressed as:
	\[
	\mathbf{C}_b^n(\hat{\psi}) = 
	\begin{bmatrix}
		\cos\hat{\psi} & -\sin\hat{\psi} \\
		\sin\hat{\psi} & \cos\hat{\psi}
	\end{bmatrix}.
	\]
		\end{itemize}

	\noindent The total loss of our MoRPI-PINN model is a weighted sum of the above three components \eqref{eq:data}, \eqref{eq:ic}, and \eqref{eq:phyloss}:
	\begin{equation}
		\mathscr{L}_{\text{total}} = \lambda_{\text{data}} \mathscr{L}_{\text{data}} + \lambda_{\text{init}} \mathscr{L}_{\text{init}} + \lambda_{\text{phys}} \mathscr{L}_{\text{phys}},
		\label{pinn_loss}
	\end{equation} 
	where $\lambda_{\text{data}}$, $\lambda_{\text{init}}$, and $\lambda_{\text{phys}}$ are hyperparameters controlling the relative importance of each loss component. These weights are tuned to ensure that both empirical measurements and physical laws are adequately represented during training of neural nework.

\subsection{Network Architecture}
\noindent The architecture of the MoRPI-PINN model for state estimation of mobile robots is designed to capture the intricate dynamics of navigation. To handle the sensors data and approximate a system of governing equations, dedicated DNN is designed. Its architecture consists of 10 hidden layers with 128 neurons each. We apply layer normalization on each layer and use the $SinTanh$ nonlinear activation function \cite{sahoomachine}. The activation function $\phi(\cdot)$ is defined as:
\begin{equation}
 \phi(z^{(\ell)}_{\dot{\imath}}) = \sin(z^{(\ell)}_{\dot{\imath}}) \cdot \tanh(z^{(\ell)}_{\dot{\imath}}),
\end{equation}

%\begin{equation}
%	\frac{\partial  \phi(z^{(\ell)})}{\partial z^{(\ell)}_{\dot{\imath}}} = \cos(z^{(\ell)}_{\dot{\imath}}) \cdot \tanh(z^{(\ell)}_{\dot{\imath}}) + \sin(z^{(\ell)}_{\dot{\imath}}) \cdot \left(1 - \tanh^2(z^{(\ell)}_{\dot{\imath}})\right).
%\end{equation}

\noindent where $ z $ is the output layer defined by: 
	\begin{equation}
		z^{(\ell)}_{\dot{\imath}} = \sum_{\dot{\jmath}=1}^{n_{\ell-1}} \omega^{(\ell)}_{\dot{\imath}\dot{\jmath}} a^{(\ell-1)}_{\dot{\jmath}} + b^{(\ell)}_{\dot{\imath}},
	\end{equation}
where \(\bm{ \theta} = \{\bm{\omega}^{[\ell]},{ \mathbf{b}}^{[\ell]}\}\), denotes the set of trainable parameters, \( \ell \) is the index of the current layer, \( L \) is the total number of layers in the neural network, and \( n_\ell \) represents the number of neurons in the \( \ell^{\text{th}} \) layer.

\begin{figure*}
	\centering
	\includegraphics[width=0.9\textwidth]{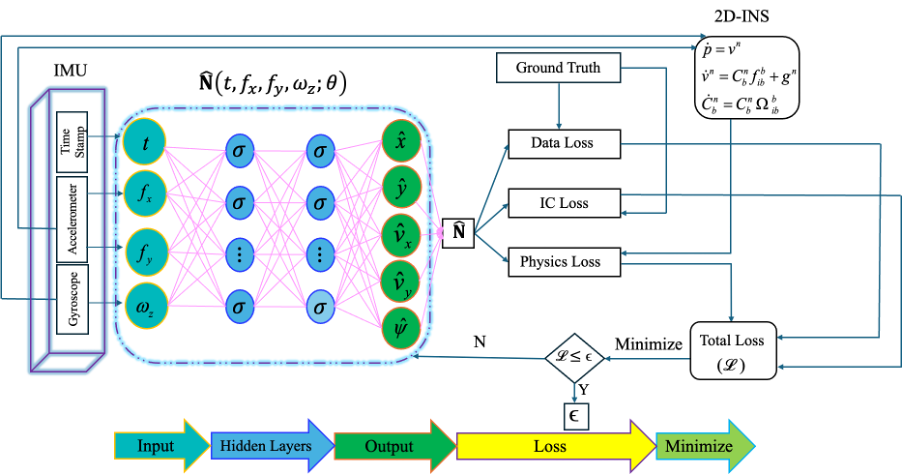}
	\caption{Illustration of the proposed MoRPI-PINN framework training process.}
	\label{PINNs}
\end{figure*}

	\noindent	Figure~\ref{PINNs} illustrates the architecture of our proposed MoRPI-PINN framework for mobile robot pure inertial navigation. The network receives time ($t$), specific force measurements from the accelerometer ($f_x$, $f_y$), and angular velocity from the gyroscope ($\omega_z$) as inputs. These are processed through a fully connected neural network to predict the robot's 2D position ($\hat{x}, \hat{y}$), velocity ($\hat{v}_x, \hat{v}_y$), and yaw angle ($\hat{\psi}$), collectively denoted as $\hat{\mathbf{N}}$. 
	
		\subsection{Training Process}
\noindent Since the PINNs approach is mesh-free, the scattered training points have been employed instead of lattice-like structured points. This provides the network more flexibility for training of complex trajectories.
	The model has been trained for $ 1,000  $ epochs using a batch size of 32 and an initial learning rate of $ 0.001 $.  To enhance generalization and prevent overfitting, dropout was applied with a rate of $ 0.2 $. A learning rate scheduler reduced the learning rate ($\eta$) by a factor of $ 0.5 $ if the validation loss plateaued for $ 50 $ epochs, enabling broad initial adjustments. Early stopping was implemented with a patience to prevent overfitting. The inertial data were segmented using overlapping windows of 10 samples with an overlap of 12 samples between successive segments to preserve temporal coherence. 

		\begin{table}[h!]
		\centering
		\caption{MoRPI-PINN hyperparameters}
		\label{tab:pinn_hyperparameters}
		\begin{tabular}{lc}
			\hline
			\textbf{Hyperparameter} & \textbf{Value} \\
			\hline
			Learning Rate & 0.001 \\
			Batch Size & 32 \\
			Epochs & 1000 \\
			Hidden Layers & 10 \\
			Neurons per Layer & 128 \\
			Dropout Rate & 0.2 \\
			\hline
		\end{tabular}
	\end{table}
		\noindent During the training, network parameters $\bm{\theta}$ are optimized using the Adam optimizer, known for its efficacy in handling complex optimization challenges. The optimizer has employed with weight decay ($ 1e-04 $) to provide $  L2 $ regularization. The estimated state ($  \hat{\mathbf{N}}(t)  $) and its derivative with respect to the trainable parameters are derived through the PINNs framework. The hyperparameters used for training the MoRPI-PINN model are summarized in Table~\ref{tab:pinn_hyperparameters}. The configuration of our hardware used for training the models is summarized in Table~\ref{tab:gpu_config}. The MoRPI-PINN training process is summarized in Algorithm 1.

	\begin{table}[h!]
	\centering
	\caption{Hardware setup: GPU specifications for model training}
	\label{tab:gpu_config}
	\begin{tabular}{ll}
		\hline
		\textbf{Parameter}       & \textbf{Details}                \\
		\hline
		GPU Model               & NVIDIA GeForce RTX 4090         \\
		CUDA Version            & 12.6                            \\
		Driver Version          & 560.35.05                       \\
		Memory Size             & 24 GB                           \\
		Tensor Cores            & 512                             \\
		Compute Capability      & 8.9                             \\
		Operating System        & macOS 15.4.1                    \\
		\hline
	\end{tabular}
\end{table}

	\begin{algorithm} \label{algo_1}
		\caption{MoRPI-PINN training process}
		\begin{algorithmic}[1]
			\State \textbf{Input:} Initial weights $\mathbf{w}_0$, biases $\mathbf{b}_0$
			\State \textbf{Given:} 
			\begin{itemize}
				\item Supervised data $\{t_i, \mathbf{x}_{\text{true}}(t_i)\}_{i=1}^{N_{\text{data}}}$
				\item Initial state $\mathbf{x}_0$ at $t=0$
				\item Collocation points $\{t_j\}_{j=1}^{N_{\text{phys}}}$
			\end{itemize}
			\State \textbf{Initialize:} Iteration counter $k \gets 0$, convergence threshold $\epsilon$
			\While{not converged}
			\State Predict: $\hat{\mathbf{x}}(t)$ using current network parameters
			\State Compute total loss: $\mathscr{L}_{\text{total}}$ according to \eqref{pinn_loss}
			\State Update network parameters using suitable optimizer
			\If{change in $\mathscr{L}_{\text{total}} < \epsilon$}
			\State \textbf{break}
			\EndIf
			\State $k \gets k + 1$
			\EndWhile
		\end{algorithmic}
	\end{algorithm}

	\section{Analysis and Results} \label{sec4}
\noindent	We begin by describing the dataset used in our analysis followed by a definition of evaluation matrices. Then, we conduct a thorough analysis of our proposed approach and compare it to the model-based baseline (2D-INS) and the data-driven baseline (MoRPINet).

%In this section, we have showcased the effectiveness of the PINNs technique through various experiments. Specifically, four cases were presented to estimate position and velocity of remote-controlled (RC) car (c.f. Figure \ref{car}) over time using inertial measurement data. All experiments were carried out in GPU with specifications listed in Table \ref{tab:gpu_config} within the Jupyter notebook environment, leveraging the PyTorch library for AD. The accuracy of the proposed algorithm is presented in multiple graphical and tabular representations. 

\subsection{Dataset}
\begin{figure} [!ht]
		\begin{center}
			\includegraphics[height=7cm,width=7cm]{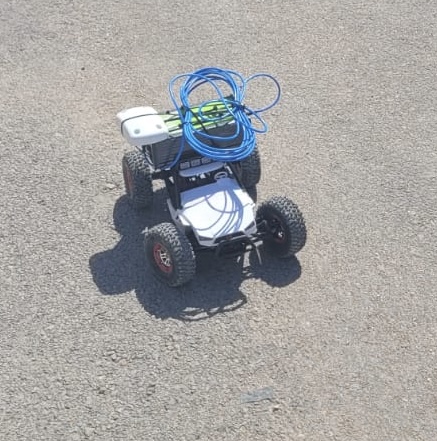}
			\caption{RC car used in the experiments mounted with RTK GNSS and IMUs.}
			\label{car}
		\end{center}
	\end{figure}

\noindent	The experimental GT data were collected from a field experiment conducted by Etzion et al. \cite{etzion2025snake}. In this experimment, a RC car, model STORM Electric 4WD Climbing Car, was employed (Figure \ref{car}). The car has dimensions of $385 \times 260 \times 205\,\text{mm}$, a wheelbase of $253\,\text{mm}$, and a tire diameter of $110\,\text{mm}$. The RC car was equipped with a \textit{Javad SIGMA-3N} RTK GNSS sensor \cite{javad_sigma}, providing high-accuracy positioning measurements with an accuracy of $10\,\text{cm}$ at a sample rate of $10\,\text{Hz}$, serving as the GT. Additionally, the RC car was mounted with \textit{Movella DOT} IMUs, capable of operating at $120\,\text{Hz}$ \cite{xsens_dot}. Further details regarding the dataset are provided in \cite{etzion2025snake}. Since the IMU operates at a frequency of $ 120 Hz $ and the RTK at $ 10 Hz $, one RTK sample corresponds to twelve IMU samples. The associated noise and bias values
of the accelerometer and gyroscope are presented in Table \ref{tab:imu}.

	\begin{table}[h!]
		\centering
		\caption{Movella DOT IMU specifications~\cite{xsens_dot}.}
		\renewcommand{\arraystretch}{1.2}
		\begin{tabular}{lcc}
			\toprule
			\textbf{Specification} & \textbf{Gyroscope} & \textbf{Accelerometer} \\
			\midrule
			Bias & 10\,$^\circ$/h & 0.03\,mg \\
			Noise Density & 0.007\,$^\circ$/s/$\sqrt{\text{Hz}}$ & 120\,$\mu$g/$\sqrt{\text{Hz}}$ \\
			\bottomrule
		\end{tabular} \label{tab:imu}
	\end{table}
	
\begin{figure}
		\begin{center}
			\includegraphics[height=5cm,width=7cm]{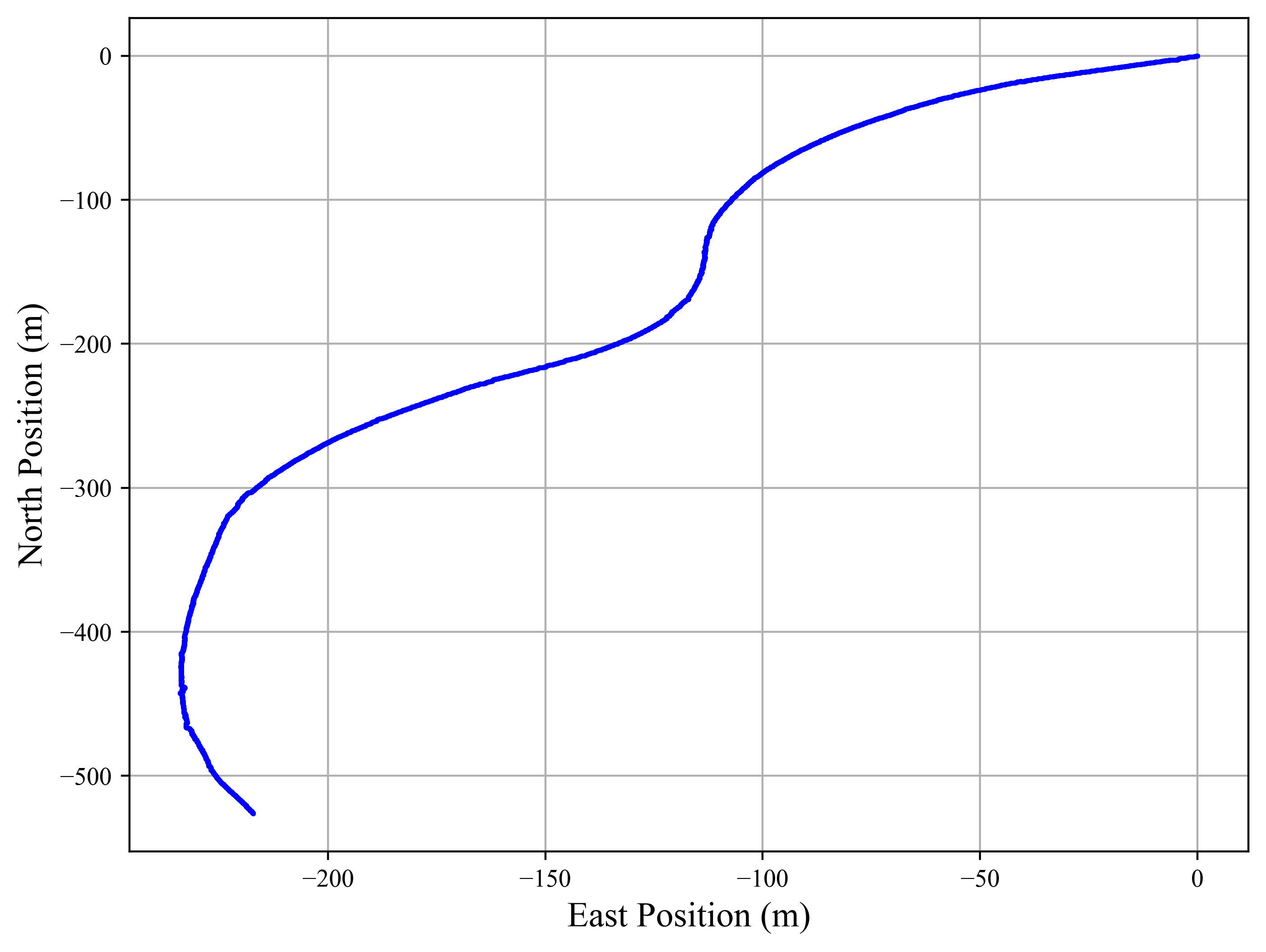}
			\caption{The single ground-truth trajectory of the training set spans 794 sec.}
			\label{Train_trajectory}
		\end{center}
	\end{figure}

\begin{figure*}[!h]
		\centering
		\subfloat[Trajectory 1\label{Test_trajectory_1}]{
			\includegraphics[width=0.48\textwidth]{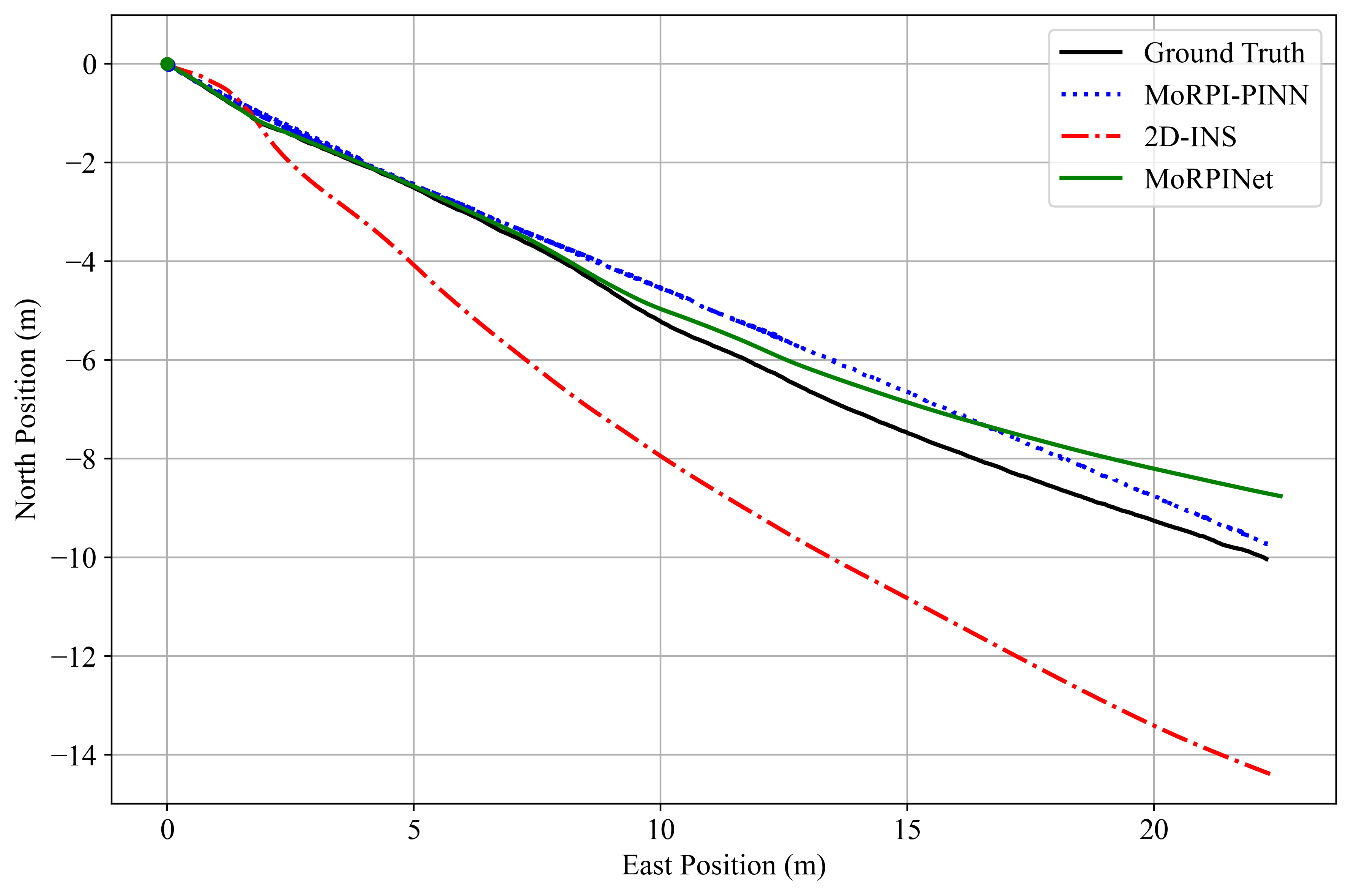}
		}
		\hfill
		\subfloat[Trajectory 2\label{Test_trajectory_2}]{
			\includegraphics[width=0.48\textwidth]{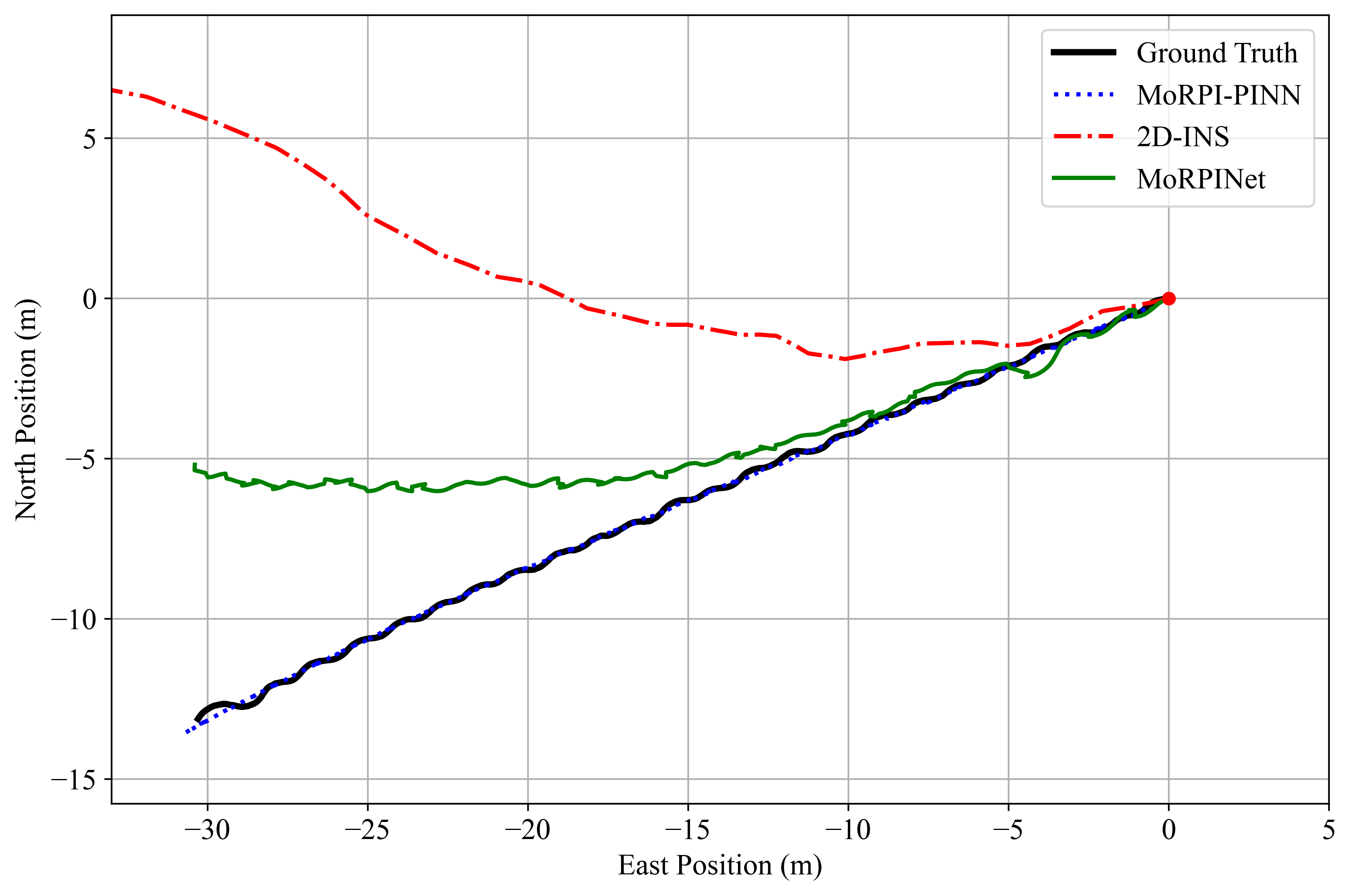}
		}
		
		\par\medskip
		
		\subfloat[Trajectory 3\label{Test_trajectory_3}]{
			\includegraphics[width=0.48\textwidth]{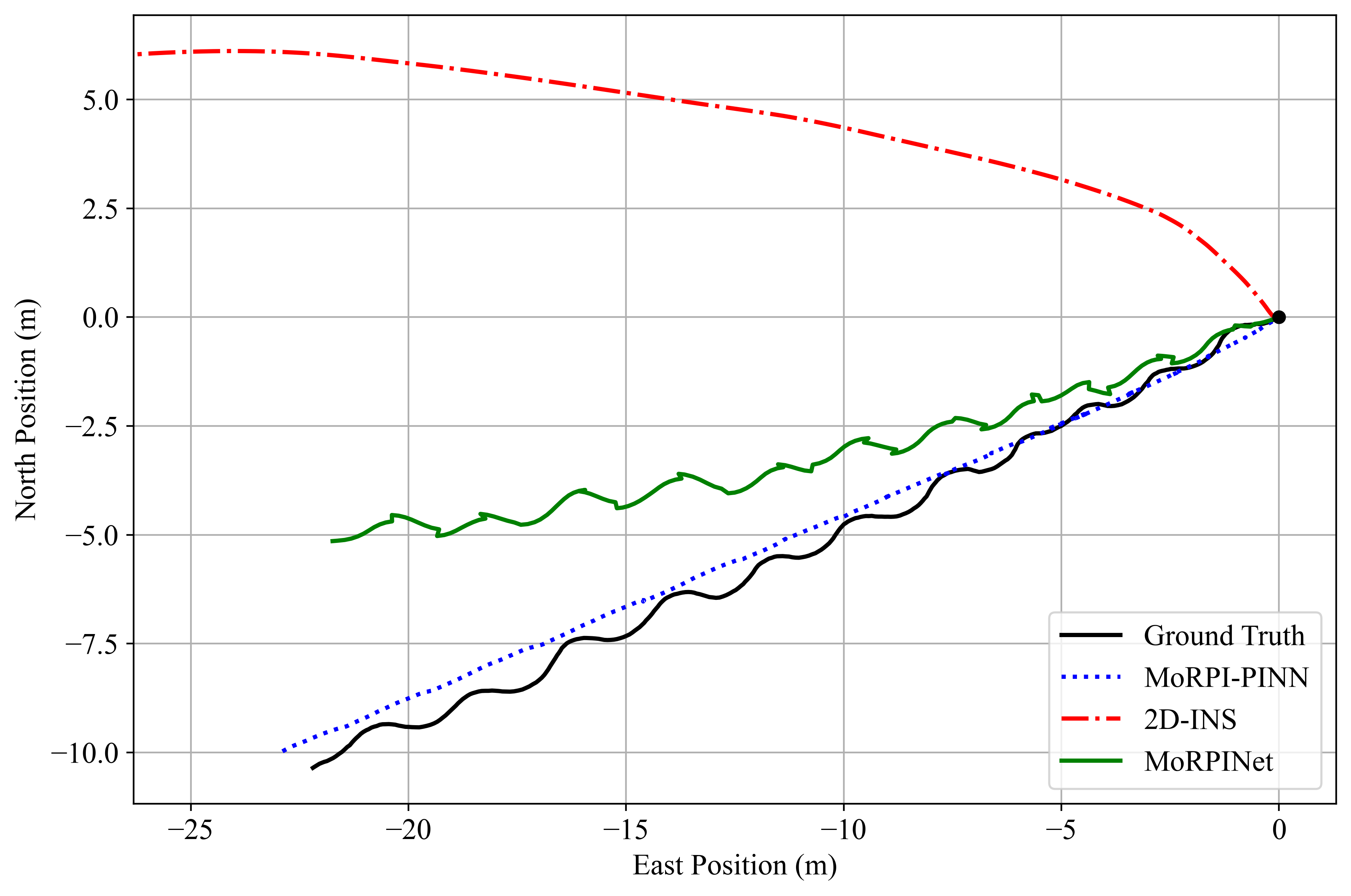}
		}
		\hfill
		\subfloat[Trajectory 4\label{Test_trajectory_4}]{
			\includegraphics[width=0.48\textwidth]{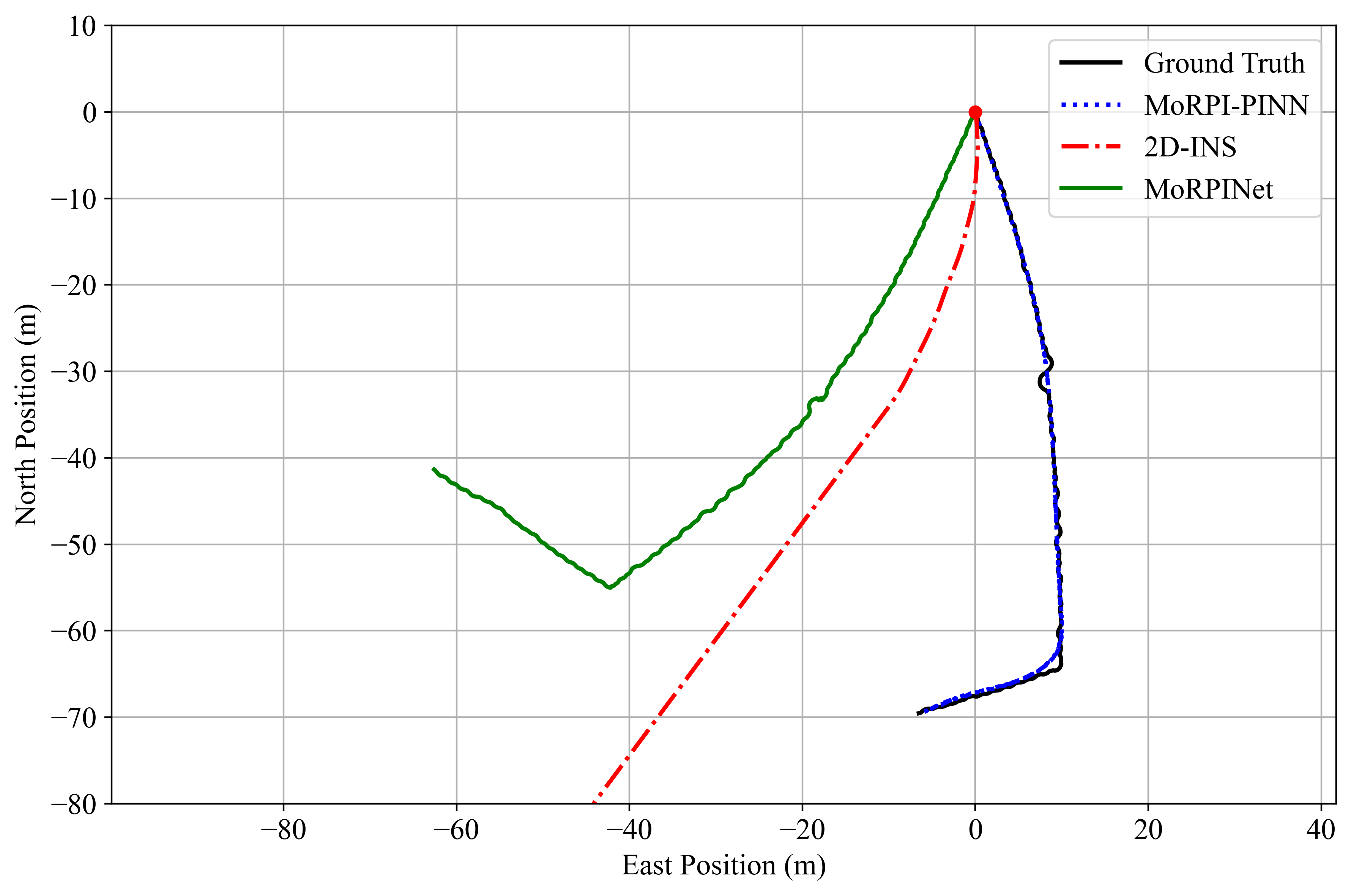}
		}
		
		\caption{Ground truth trajectory of the test dataset with the estimated trajectory from the model-based baseline (2D-INS), data-driven baseline (MoRPINet), and our MoRPI-PINN method.}
		\label{test_trajectories}
	\end{figure*}

	\noindent The training trajectory, shown in Figure~\ref{Train_trajectory}, spans a duration of 794 sec and serves as the primary data for the training. The testing dataset comprises four trajectories from the same experimental study~\cite{etzion2025snake}, collectively covering approximately 25~m. These trajectories include both straight-line and periodic motion patterns. The four test trajectories, depicted in Figure~\ref{test_trajectories}, were recorded over 37, 46, 39, and 70 sec, respectively, totaling 192 sec of inertial data. The same dataset is used for evaluating the baseline methods and our proposed MoRPI-PINN framework. 
	
	\noindent A total of 200 initial condition points are incorporated into training. These are synthetically generated by duplicating the first observed sensor readings from the training trajectory (Figure~\ref{Train_trajectory}) to provide stable initialization across collocation samples.
	Moreover, prior to training and testing, the data underwent the following preprocessing pipeline:
	\begin{enumerate}
		\item The first 60 sec of data were removed for initial stabilization.
		\item A new time stamp starting from 0 sec was asssigned according to the sampling frequency of the corresponding sensors.
		\item Each axis was normalized independently using its respective direction.
	\end{enumerate}

	\subsection{Evaluation Metrics}
	\label{sec:evaluation}
	\noindent	For our analysis, we use the following evaluation metrics:
	
	%Following the training of the MoRPI-PINN using the above discussed dataset, its performance is evaluated using various global error metrics computed on test data for position (\(x, y\)) and velocity (\(v_x, v_y\)). These metrics are compared with the model's internal loss components such as physics loss (\(\mathscr{L}_{\text{phys}}\)), data loss (\(\mathscr{L}_{\text{data}}\)), and initial condition loss (\(\mathscr{L}_{\text{init}}\)) in order to assess accuracy, and robustness for inertial navigation. The error metrics can be represented as:
	
	\begin{itemize}
		\item \textbf{Absolute Trajectory Error (ATE):}
		% It is defined as the Euclidean distance between predicted and actual positions along the trajectory:
		\begin{equation}
			\text{ATE}(\textbf{p}_i, \hat{\textbf{p}}_i) = \sqrt{\frac{1}{N} \sum_{i=1}^N (\textbf{p}_i - \hat{\textbf{p}}_i)^2 },
			\label{eq:ate}
		\end{equation}
		where $\textbf{p}_i$ and $ \hat{\textbf{p}}_i $ are the GT position vector,  and the predicted position vector, respectively, \( i \) is the time index, and \( N \) is the total number samples. 
		
		%We have also evaluated mean ATE, standard deviation of ATE, and maximum ATE to quantify overall positional accuracy, critical for navigation performance.

%		\item \textbf{Velocity Root Squared Error (VRSE):}
%		\begin{equation}
%			\text{VRSE}( \textbf{v}_i, \hat{\textbf{v}}_i) = \sqrt{  ( \textbf{v}_i - \hat{\textbf{v}}_i)^2  }.
%			\label{eq:vmse}
%		\end{equation}
%		
		
		\item \textbf{Velocity Root Mean Squared Error (VRMSE):} 
		\begin{equation}
			\text{VRMSE}( \textbf{v}_i, \hat{\textbf{v}}_i) = \sqrt{\frac{1}{N} \sum_{i=1}^N (\textbf{v}_i - \hat{\textbf{v}}_i)^2}.
			\label{eq:vrmse}
		\end{equation}
	where $\textbf{v}_i$ and $ \hat{\textbf{v}}_i $ are the GT velocity vector, and the predicted velocity vector, respectively, \( i \) is the time index, and \( N \) is the total number samples.

%		%VRMSE is particularly useful for assessing the dispersion of errors, with higher sensitivity to outliers. 
		
		\item \textbf{Normalized Velocity Root Mean Squared Error (NVRMSE):} 
		\begin{equation}
			\text{NVRMSE} = \frac{\text{VRMSE}}{\| \textbf{v}_{\max} \| - \|\textbf{ v}_{\min} \|},
			\label{eq:nvrme}
		\end{equation}
		
		where \(\textbf{v}_{\max}\) and \(\textbf{v}_{\min}\) 
		represent the maximum and minimum values of the GT velocity vector across the entire training dataset. NVRMSE scales VRMSE by the range of the GT data, enabling cross dataset comparisons and unit-free error interpretation. 
		%NVRMSE contextualizes VRMSE by making it easier to assess whether the magnitude of the errors is significant in the context of the application and meeting specific navigation precision requirements.\\
	\end{itemize}
	
	\subsection{Performance Analysis}
	\noindent	Before diving into the results, we note that for mobile robot positioning performance we compare our approach to both baseline methods.
 	However, for velocity comparison, only the 2D-INS model is considered, as MoRPINet does not provide velocity information. The 2D-INS were computed using \eqref{eq:position}--\eqref{eq_ins}, while MoRPINet was evaluated under the same conditions for comparative analysis. To improve baseline accuracy, zero-order calibration was applied to the accelerometer and gyroscope data by leveraging the initial stationary segment of each trajectory. In contrast, the proposed MoRPI-PINN model and MoRPINet used raw sensor data, in order to demonstrate the  capability to operate effectively in noisy conditions. 
 	
 		\noindent All three approaches were evaluated on the test dataset which consist of four different trajectories: straight-line trajectory (Figure~\ref{Test_trajectory_1}), periodic trajectories (Figures~\ref{Test_trajectory_2}--\ref{Test_trajectory_3}), and the L-shaped path (Figure~\ref{Test_trajectory_4}).

	\subsubsection{2D-INS}
	\noindent The 2D-INS solution was evaluated, as it is one of the most widely adopted baseline models in navigation systems. As anticipated, it produced significant errors. For position estimation, the model yields an average ATE of 14.3\,m across the test dataset.
	
	%various trajectories, including the straight-line trajectory (Figure~\ref{Test_trajectory_1}), periodic trajectories (Figures~\ref{Test_trajectory_2}--\ref{Test_trajectory_3}), and the L-shaped path (Figure~\ref{Test_trajectory_4}). 
	
	\noindent Furthermore, the accuracy of 2D-INS model in velocity estimation is notably poor, with NVRMSE reaching 1077\% for $v_x$ and 1211\% for $v_y$ in the straight-line trajectory. In the periodic trajectories, the errors remain high, with 51\% and 1060\% for $v_x$, and 37\% and 2149\% for $v_y$ in Trajectories 2 and 3, respectively. The challenges posed by the L-shaped trajectory further amplify these errors, with NVRMSEs reaching 277\% for $v_x$ and 175\% for $v_y$. These results highlight the limitations of the model-based dead-reckoning approach, in handling long durations  under dynamic motion, and sharp turns.

\subsubsection{MoRPINet}
\noindent The data-driven MoRPINet serves as a strong baseline method for navigation, offering a significant improvement in position estimation accuracy over traditional 2D-INS. The data-driven MoRPINet achieves accuracy of 85\%, with an average ATE of 5.7\,m across all trajectories. This demonstrates the potential of deep learning models to correct inertial drift by capturing data patterns directly from sensor inputs. 

\noindent Moreover, despite its gains in position estimation, it still struggles in scenarios involving sudden directional changes, where its predictions deviate more significantly. 

\subsubsection{MoRPI-PINN}
\noindent Achieves comparatively low ATEs across all trajectories, with an average ATE of 0.8\,m, resulting in 94\% improvement over 2D-INS and 85\% improvement over MoRPINet. 

\noindent In terms of velocity estimation, MoRPI-PINN demonstrates consistently low NVRMSEs. For the straight-line trajectory, it achieves 54\% in $v_x$ and 25\% in $v_y$. In the periodic motion trajectories, NVRMSEs are 34\% ($v_x$) and 49\% ($v_y$) for Trajectory 2, and 37\% ($v_x$) and 35\% ($v_y$) for Trajectory 3. For the L-shaped trajectory, the model further reduces the errors to 20\% in $v_x$ and 24\% in $v_y$. 

\noindent These results highlight the robustness and generalizability of MoRPI-PINN across a diverse set of trajectories. It demonstrates the advantages of integrating physics of 2D-INS equations of motion for accurate mobile robot navigation under challenging dynamics.
	\begin{table*} [h!]
	\centering
	\caption{Absolute Trajectory Error(m) over four trajectories with average and improvement}
	\renewcommand{\arraystretch}{1.2}
	\begin{tabularx}{\textwidth}{l*{4}{>{\centering\arraybackslash}X} >{\centering\arraybackslash}X >{\centering\arraybackslash}X}
		\toprule
		\multirow{2}{*}{\textbf{Approach}} & \multicolumn{4}{c}{\textbf{ATE [m]}} & \multirow{2}{*}{\textbf{Average}} & \multirow{2}{*}{\textbf{Improvement }} \\
		\cmidrule(lr){2-5}
		& \textbf{Trajectory 1} & \textbf{Trajectory 2} & \textbf{Trajectory 3} & \textbf{Trajectory 4} &  &  \\
		\midrule
		2D-INS (model-based baseline) & 11.2 & 7.1 & 11.3 & 27.9 & 14.3 & 94\% \\
		MoRPINet (data-driven baseline) & 5.8 & 5.1 & 5.4 & 6.6 & 5.7 & 85\% \\
		\rowcolor{gray!20}
		MoRPI-PINN (ours) & 1.1 & 0.9 & 0.9 & 0.4 & 0.8 &-- \\
		\bottomrule
	\end{tabularx}
	\label{tab:ate_comparison}
\end{table*}
	\subsection{Summary}
\begin{figure*} [h!]
		\centering
		\subfloat[]{%
			\includegraphics[width=0.48\textwidth]{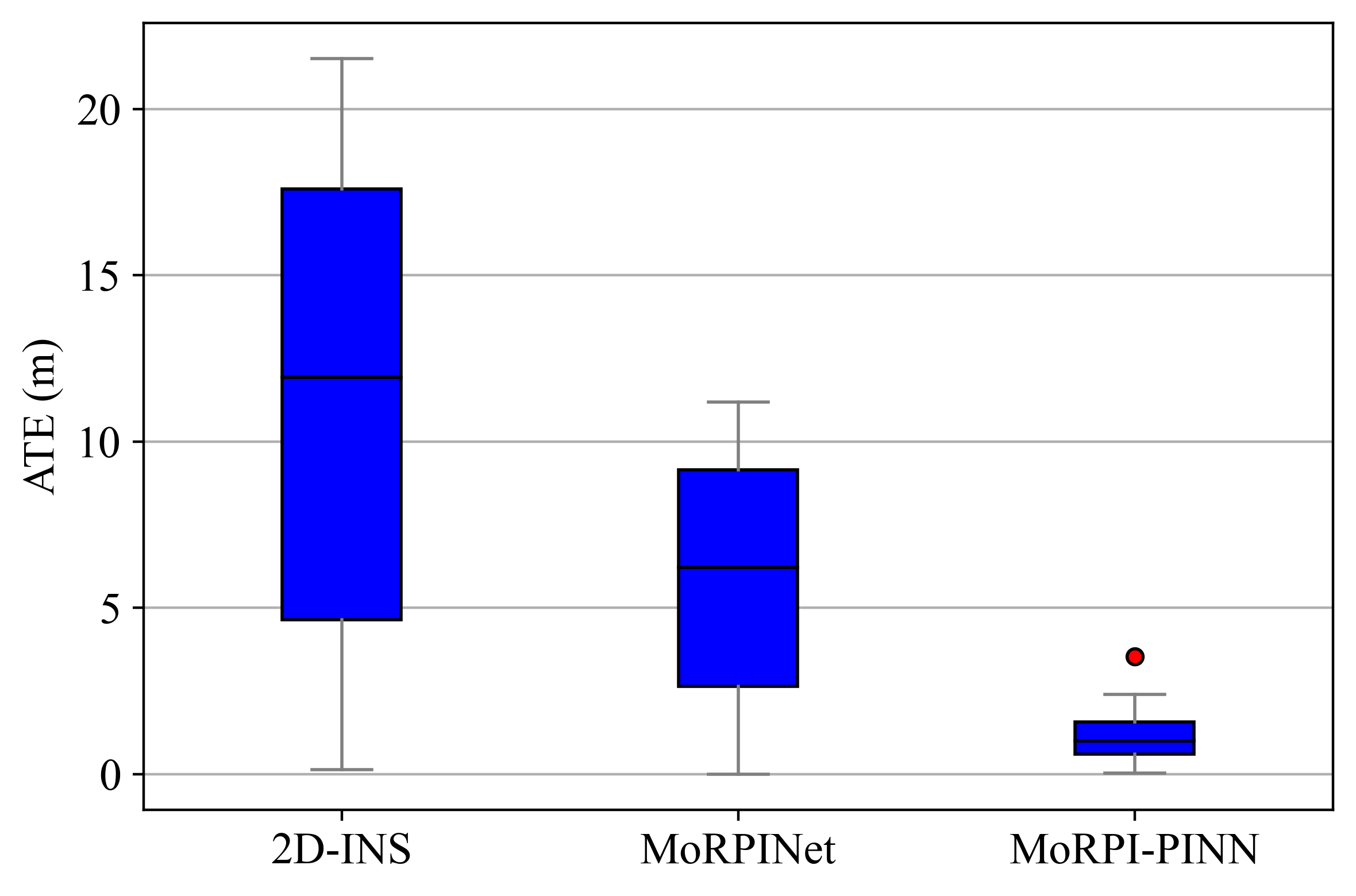}
			\label{fig:pic1}
		}
		\hfill
		\subfloat[]{%
			\includegraphics[width=0.48\textwidth]{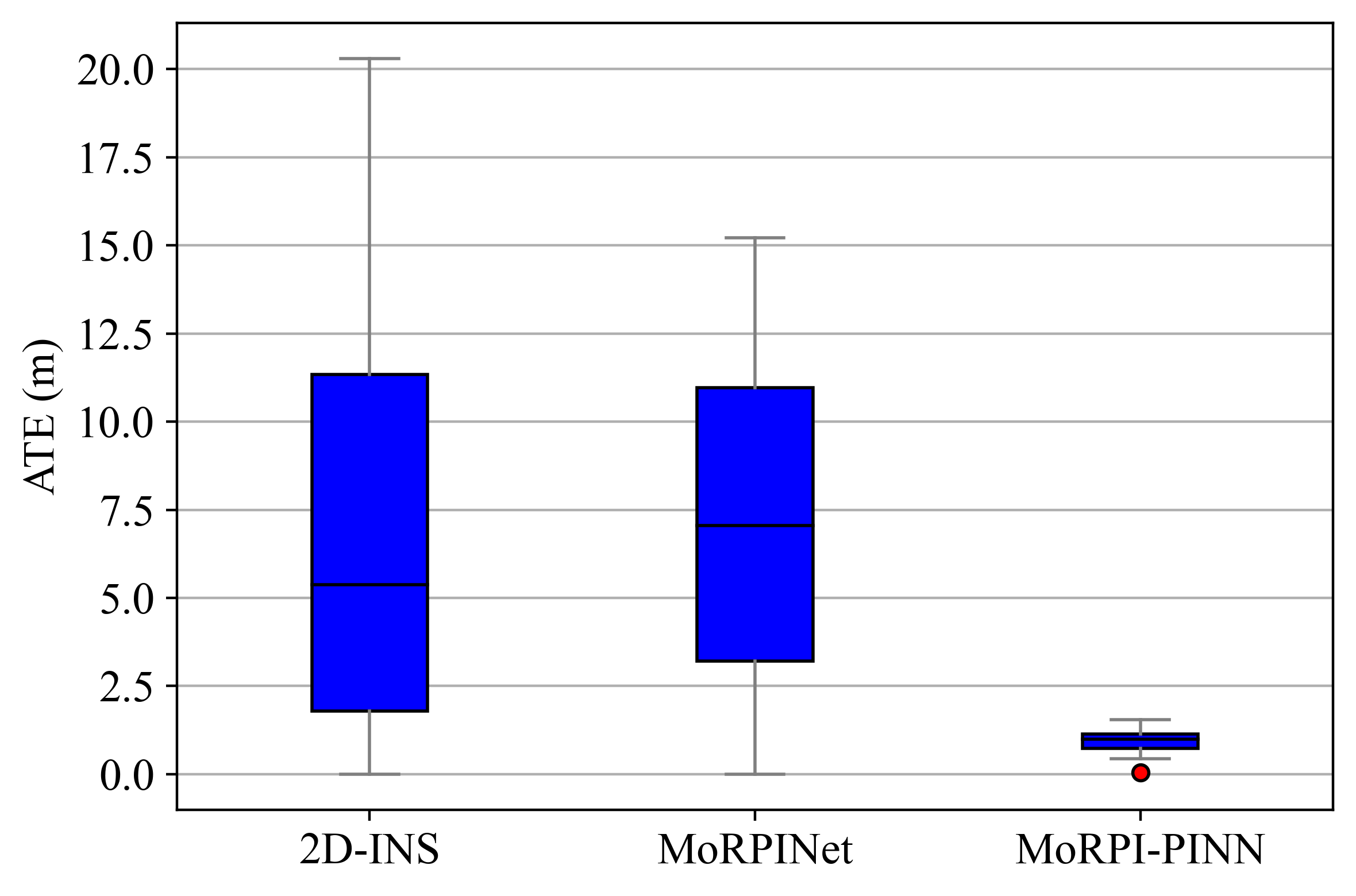}
			\label{fig:pic2}
		}
		
		\par\medskip
		
		\subfloat[]{%
			\includegraphics[width=0.48\textwidth]{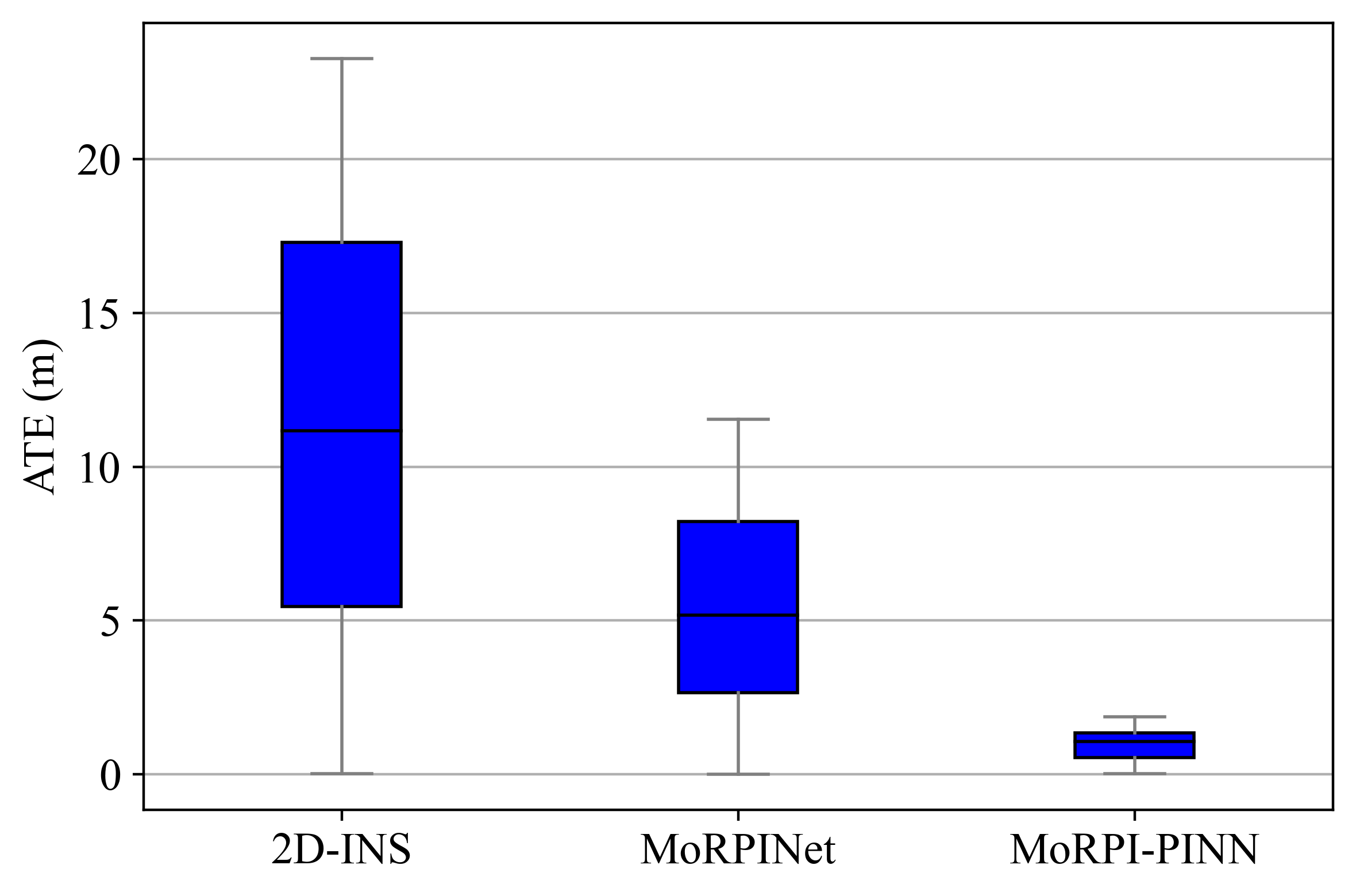}
			\label{fig:pic3}
		}
		\hfill
		\subfloat[]{%
			\includegraphics[width=0.48\textwidth]{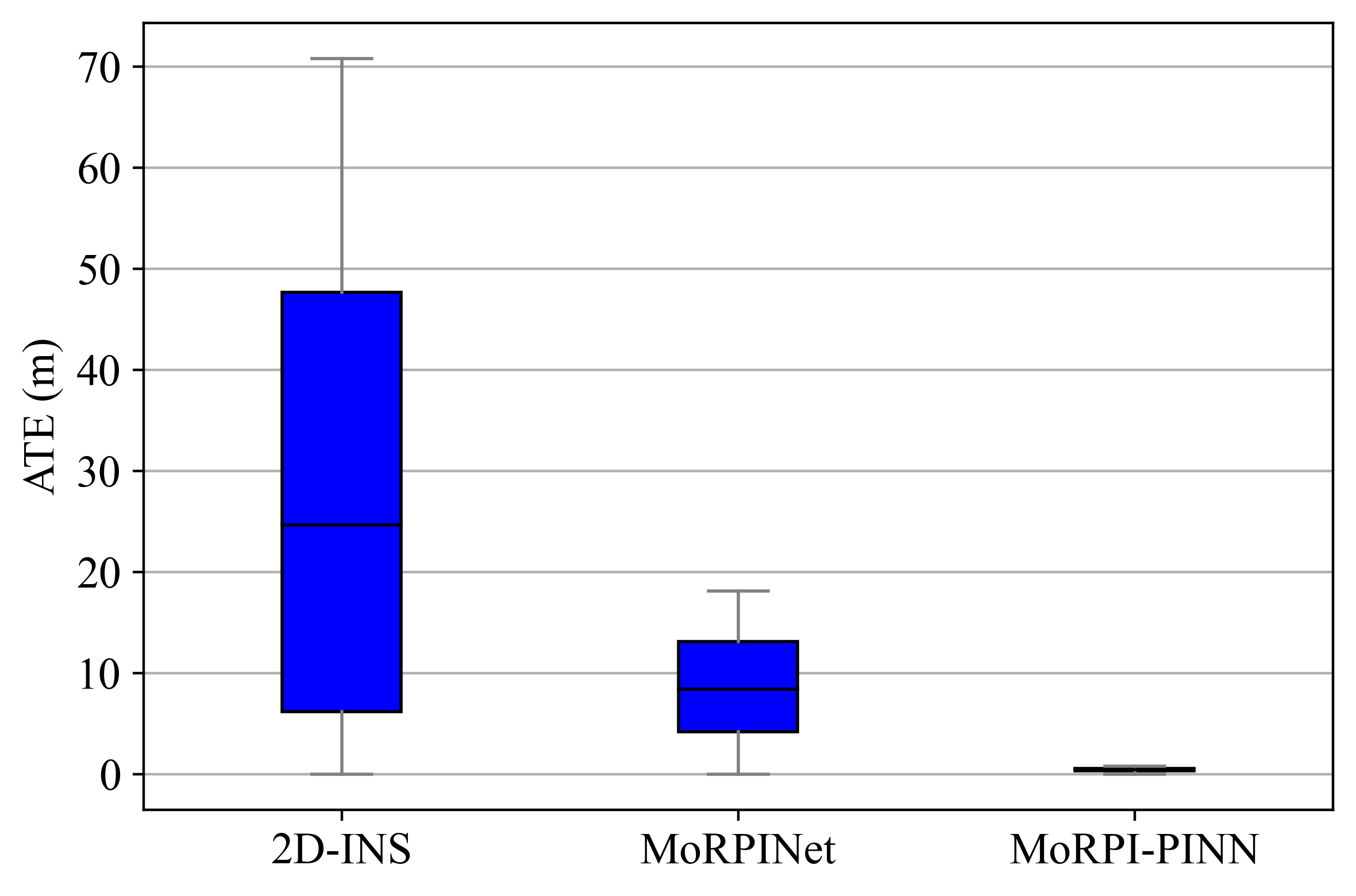}
			\label{fig:pic4}
		}
		
		\caption{ATE across the four test trajectories for the model-based baseline (2D-INS), data-driven baseline (MoRPINet), and our MoRPI-PINN method.}
		\label{ate}
	\end{figure*}

\noindent	Across all evaluated scenarios, MoRPI-PINN framework consistently and significantly outperforms the baseline models. As shown in Table~\ref{tab:ate_comparison}, MoRPI-PINN framework improved the ATE of the 2D-INS by 94\% and MoRPINet by 85\%. The model also demonstrates excellent dead-reckoning accuracy, with a maximum ATE of 1.1\,m, across all testing trajectory as shown in Figure \ref{ate}. The maximum errors remain low due to the constraints imposed by physics-informed learning in the term of the physics loss ($\mathcal{L}_{\text{phys}}$) from 2D-INS equation. In contrast to traditional 2D-INS algorithms, where errors accumulate over time, MoRPI-PINN exhibits only occasional larger deviations. So model effectively mitigates long-term drift, a critical limitation in traditional inertial navigation approach. 

\noindent	Furthermore, low initial condition loss ($\mathcal{L}_{\text{init}} \approx 10^{-3}$) ensures accurate trajectory initialization ($x(0), y(0) \approx 0$), helping to reduce ATE at the beginning of the trajectory.  Input normalization enables consistent performance across a range of accelerometer ($f_x, f_y$) and gyroscope ($\omega_z$) data and reducing errors in diverse test scenarios. Here, the physics-informed loss ($\mathcal{L}_{\text{phys}}$), again plays a key role for capturing the underlying dynamics. This regularization helps prevent overfitting to noisy measurements; a common limitation of purely data-driven neural networks.

%\noindent	Figure \ref{fig:vel_comparison_all}, represents visual comparison of the errors for estimating velocities components  $ v_x $ and $ v_y $. The MoRPI-PINN model demonstrates low velocity errors, with $\text{VRMSE}_{v_x} \approx 0.37\, \text{m/s}$, $\text{VRMSE}_{v_y} \approx 0.35\, \text{m/s}$.   In contrast, the traditional 2D-INS exhibits significantly higher errors, with $\text{VRMSE}_{v_x} \approx 6.00\, \text{m/s}$, $\text{VRMSE}_{v_y} \approx 8.02\, \text{m/s}$, respectively. 

\noindent	These results emphasize the potential of physics-informed learning to significantly enhance inertial navigation, especially in scenarios where sensor calibration is impractical or where the motion deviates from simple linear dynamics.
	
%		\begin{figure*}[h!]
%		\centering
%		\subfloat[]{%
%			\includegraphics[width=0.48\textwidth]{vel_comparison_1}
%			\label{fig:vel_1}
%		}
%		\hfill
%		\subfloat[]{%
%			\includegraphics[width=0.48\textwidth]{vel_comparison_2}
%			\label{fig:vel_2}
%		}
%		
%		\par\medskip
%		
%		\subfloat[]{%
%			\includegraphics[width=0.48\textwidth]{vel_comparison_3}
%			\label{fig:vel_3}
%		}
%		\hfill
%		\subfloat[]{%
%			\includegraphics[width=0.48\textwidth]{vel_comparison_4}
%			\label{fig:vel_4}
%		}
%		
%		\caption{Comparisons of VRSE across four scenarios.}
%		\label{fig:vel_comparison_all}
%	\end{figure*}
%	

%	\begin{table}[h]
%		\centering
%		\caption{Comparison of velocity root mean squared error over four trajectories (in m/s)}
%		\begin{tabular}{lcccc}
%			\toprule
%			\textbf{Trajectory} & \textbf{$ v_x $\_PINN} & \textbf{$ v_x $\_2D-INS} & \textbf{$ v_y $\_PINN} & \textbf{$ v_y $\_2D-INS} \\
%			\midrule
%			Trajectory 1 & \textbf{\textcolor{blue}{0.42}} & 8.43 & 0.31 & 15.02 \\
%			Trajectory 2 & 0.37 & 0.55 & \textbf{\textcolor{blue}{0.36}} & 0.27 \\
%			Trajectory 3 & 0.38 & 10.65 & 0.28 & 17.51 \\
%			Trajectory 4 & 0.32 & 4.38 & 0.45 & 3.31 \\
%			\midrule
%			\rowcolor{gray!20}
%			\textbf{Average} & 0.37 & 6.00 & 0.35 & 8.02 \\
%			\bottomrule
%		\end{tabular}
%		\label{tab:rmse}
%	\end{table}
%	

	\section{Conclusion} \label{sec5}
	\noindent  In real-world scenarios, GNSS-denied environments and poor lighting conditions result in pure inertial navigation and thus rapid drift in position solution. To mitigate this drift, we introduced the MoRPI-PINN framework. Our approach integrates 2D-INS governing equations into the neural network training process. It employs compound loss combining data loss, physical loss, and initial condition loss. This design enables robust learning even in the presence of sensor error terms and noise.
	
	\noindent We assessed MoRPI-PINN across four real-world trajectories and compared it against two existing model-based methods: (1) a traditional numerical 2D-INS solution, and (2) the data-driven MoRPINet approach. MoRPI-PINN achieved a position error of just 0.8 m, significantly outperforming 2D-INS (14.3\,m) and MoRPINet (5.7\,m). This corresponds to 94\% improvement over the 2D-INS solution and 85\% improvement over the MoRPINet baseline. Moreover, MoRPI-PINN demonstrated robust performance in unstructured and low-calibration environments, where conventional methods typically experience drift and bias accumulation.
	These results demonstrate that MoRPI-PINN is a robust solution for mobile robot state estimation in GNSS denied conditions. Its ability to simultaneously leverage physical constraints and data-driven learning allows it to overcome the limitations of both purely analytical and purely data-based methods. A potential limitation is that MoRPI-PINN may exhibit slightly higher ATE in data-rich scenarios, as strict adherence to physics may reduce adaptability to rapid pattern changes in the data. 
	
	\noindent Overall, the proposed model is well-suited to a wide range of real-world mobile robot applications, including manufacturing and logistics, security and surveillance, delivery services and infrastructure inspection. MoRPI-PINN is lightweight, yet has an effective architecture making it ideal for deployment on embedded platforms, enabling real-time navigation in adversarial environments. Thus, MoRPI-PINN effectively bridges the limitations of pure inertial navigation, and enables seamless robot navigation using inertial sensor data during short time-periods.

	\section*{Acknowledgement}
\noindent A. K. S.  was supported by the Maurice Hatter Foundation.
%
%%\section*{Funding}
%%No funding has received for preparation of this manuscript. 
%%
%%\section*{Data Availability}
%%All recorded data used for our evaluations are publicly available at https://github.com/ansfl/MoRPINet.
%%
%%\section*{Declaration}
%%The authors have no conflicts of interest to declare.
%%
%
%
%
%
\bibliography{New_IEEEtran}
\bibliographystyle{IEEEtran1}
\vspace*{-4em}
\end{document}